# Learning Semantic Representations for the Phrase Translation Model

**Jianfeng Gao, Xiaodong He, Wen-tau Yih, Li Deng**
Microsoft Research, Redmond
Washington 98052, USA
{jfgao, xiaohe, scottyi, deng}$microsoft.com

## Abstract

This paper presents a novel semantic-based phrase translation model. A pair of source and target phrases are projected into continuous-valued vector representations in a low-dimensional latent semantic space, where their translation score is computed by the distance between the pair in this new space. The projection is performed by a multi-layer neural network whose weights are learned on parallel training data. The learning is aimed to directly optimize the quality of end-to-end machine translation results. Experimental evaluation has been performed on two Europarl translation tasks, English-French and German-English. The results show that the new semantic-based phrase translation model significantly improves the performance of a state-of-the-art phrase-based statistical machine translation system, leading to a gain of 0.7-1.0 BLEU points.

## 1. Introduction

The phrase translation model, also known as the *phrase table*, is one of the core components of phrase-based statistical machine translation (SMT) systems. The most common method of constructing the phrase table takes a two-phase approach [17]. First, the bilingual phrase pairs are extracted heuristically from an automatically word-aligned training data. The second phase, which is the focus of this paper, is parameter estimation where each phrase pair is assigned with some scores that are estimated based on counting these phrases using the same word-aligned training data. Phrase-based SMT systems have achieved state-of-the-art performance largely due to the fact that long phrases, rather than single words, are used as translation units so that useful context information can be captured in selecting translations. However, longer phrases occur less often in training data, leading to a severe data sparseness problem in parameter estimation. There has been a plethora of research reported in the literature on improving parameter estimation for the phrase translation model [e.g., 6, 8, 10, 25].

This paper revisits the problem of scoring a phrase translation pair by developing a novel, Semantic-based Phrase Translation Model (SPTM). The translation score of a phrase pair in this model is computed as follows. First, we represent each phrase as a bag-of-words vector, called *word vector* henceforth. We then project the word vector, in either the source language or the target language, into a respective continuous feature vector in a common low-dimensional latent semantic space that is intended to be language independent. The projection is performed by a multi-layer neural network. The projected feature vector forms the *semantic representation* of a phrase. Finally, the translation score of a source-target phrase pair is computed by the distance between their feature vectors.

The main motivation behind the SPTM is to alleviate the data sparseness problem associated with the traditional counting-based methods by grouping phrases with a similar meaning across different languages. In this model, semantically related phrases, in both the source and the target languages, would have similar (close) feature vectors in the semantic space. Since the translation score is a smooth function of these feature vectors, a small change in semantics (e.g., the phrases that differ only in morphological forms) should only lead to a small change in the translation score.

The primary research task in developing the SPTM is learning the semantic representation of a phrase that is effective for SMT. Motivated by recent studies on continuous-space language models (LM) [3, 19], we use a neural network to project a word vector to a feature vector. Ideally, the projection would discover those latent semantic features that are useful to differentiate *good* translations from *bad* ones, for a given source phrase. However, there is no training data with explicit annotation on the quality of phrase translations. The phrase translation pairs are *hidden* in the parallel source-target sentence pairs, which are used to train the traditional translation models. The quality of a phrase translation can only be judged implicitly through the translation quality of the sentences, as measured by BLEU, which contain the phrase pair. In order to overcome this challenge and let the BLEU metric guide the projection learning, we propose a new method to learn the parameters of a neural network. This new method automatically forces the feature vector of a source phrase to be closer to the feature vectors of its candidate translations that lead to a better BLEU score, if these translations are selected by an SMT decoder to produce final, sentence-level translations. The new learning method makes use of the L-BFGS algorithm and the expected BLEU as the objective function defined on N-best lists.

To the best of our knowledge, the SPTM proposed in this paper is the first continuous-space phrase translation model that is shown to lead to significant improvement over a standard phrase-based SMT system (to be detailed in Section 6). Like the traditional phrase translation model, the translation score of each bilingual phrase pair is modeled *explicitly* in our model. However, instead of estimating the phrase translation score on aligned parallel data, our model intends to capture the semantic similarity between a source phrase and its paired target phrase by projecting them into a common, latent semantic space that is language independent.

## 2. Related Work

Latent Semantic Analysis (LSA) [5], originally designed for information retrieval (IR), is arguably the earliest continuous semantic model. Unlike LSA which is a linear projection model, generative topic models, such as Probabilistic LSA [13] and Latent Dirichlet Allocation (LDA) [4] give a clear probabilistic interpretation of the semantic representation. In contrast, recent work on continuous space language models, e.g., the feed-forward neural network language model (NNLM) [3] and the recurrent neural network language model (RNNLM) [19, 2], provide a different kind of latent semantic representation. Because these latent semantic models are developed for mono-lingual settings, they are not directly applicable to translation. As a result, variants of such models for cross-lingual scenarios have been proposed [7, 9, 22, 26] where documents in different languages are projected into the shared latent concept space. In principle, a phrase translation table can be derived using these cross-lingual models, although decoupling the derivation from the SMT training often results in suboptimal performance.

Despite the success of latent semantic representations in various applications, there is, however, much less work on continuous-space translation models. The only exception we are aware of is the work of continuous space *n*-gram translation models [23, 24], where the feed-forward NNLM is extended to represent translation probabilities. However, these earlier studies focused on the so-called *n*-gram translation models, where the translation probability of a phrase or a sentence is decomposed as a product of word *n*-gram probabilities of the same form as that in a standard *n*-gram LM. Therefore, it is not clear how their approaches can be applied to the phrase translation model, which is much more widely used in modern SMT systems. In contrast, our model learns jointly the representations of a phrase in the source language as well as its translation in the target language.

There has been much recent research on improving the phrase table [6, 8, 10, 18, 25]. Among them, [8] is most relevant to the work described in this paper. They estimate phrase translation probabilities using a discriminative training method under the N-best reranking framework of SMT and an N-best list based expected BLEU as the objective function. In this study we use the same objective function to learn the semantic representations of phrases, integrating the strengths associated with both of these earlier studies.

## 3. The Log-Linear Model for SMT

Phrase-based SMT is based on a log-linear model which requires learning a mapping between inputs $F \in \mathcal{F}$ to outputs $E \in \mathcal{E}$. We are given

- Training samples $(F_i, E_i)$ for $i = 1 \ldots N$, where each source sentence $F_i$ is paired with a reference translation in target language $E_i$;
- A procedure GEN to generate a list of N-best candidates $\text{GEN}(F_i)$ for an input $F_i$, where GEN in this study is the baseline phrase-based SMT system, i.e., a reimplementation of the Moses system [15] that does not use the SPTM, and each $E \in \text{GEN}(F_i)$ is labeled by the sentence-level BLEU score [10], denoted by $\text{sBleu}(E_i, E)$, which meaures the quality of $E$ with respect to its reference translation $E_i$;
- A vector of features $\mathbf{h} \in \mathbb{R}^M$ that maps each $(F_i, E)$ to a vector of feature values; and
- A parameter vector $\boldsymbol{\lambda} \in \mathbb{R}^M$, which assigns a real-valued weight to each feature.

SMT involves hidden-variable models such that a hidden variable $A$ is assumed to be constructed during the process of generating $E$. In the phrase-based SMT, $A$ consists of a segmentation of the source and target sentences into phrases and an alignment between source and target phrases.

The components $\text{GEN}(.)$, $\mathbf{h}$ and $\boldsymbol{\lambda}$ define a log-linear model that maps $F_i$ to an output as follows:

$$E^* = \underset{(E,A) \in \text{GEN}(F_i)}{\text{argmax}} \boldsymbol{\lambda}^\text{T} \mathbf{h}(F_i, E, A) \tag{1}$$

which states that given $\boldsymbol{\lambda}$ and $\mathbf{h}$, argmax returns the highest scoring translation $E^*$, maximizing over correspondences $A$. In phrase translation models, computing the argmax exactly is intractable, so it is performed approximatedly by beam search. We assume that every translation candidate is always coupled with a corresponding $A$, called *Viterbi derivation*, generated by (1).

## 4. A Semantic-Based Phrase Translation Model (SPTM)

The architecture of the SPTM is shown in Figure 1, where for each pair of source and target phrases $(f_i, e_j)$ in a source-target sentence pair, we first project them into feature vectors $\mathbf{y}_{f_i}$ and $\mathbf{y}_{e_j}$ in a latent semantic space via a neural network, and then compute the translation score, $\text{score}(f_i, e_j)$, by the distance of their feature vectors in that space.

We start with a bag-of-words representation of a phrase $\mathbf{x} \in \mathbb{R}^d$, where $\mathbf{x}$ is a word vector and $d$ is the size of the vocabulary consisting of words in both source and target languages. We then learn to project $\mathbf{x}$ to a low-dimensional semantic space $\mathbb{R}^k$: $\phi(\mathbf{x}): \mathbb{R}^d \to \mathbb{R}^k$. The projection is performed using a fully connected neural network with one hidden layer and tanh activation functions. Let $\mathbf{W}_1$ be the projection matrix from the input layer to the hidden layer and $\mathbf{W}_2$ the projection matrix from the hidden layer to the output layer, we have

$$\mathbf{y} \equiv \phi(\mathbf{x}) = \tanh\left(\mathbf{W}_2^\text{T}\left(\tanh(\mathbf{W}_1^\text{T}\mathbf{x})\right)\right) \tag{2}$$

The translation score of a source phrase *f* and a target phrase *e* can be measured as the similarity (or distance) between their feature vectors. We choose the dot product as the similarity function[1]:

$$\text{score}(f, e) \equiv \text{sim}_{\boldsymbol{\theta}}(\mathbf{x}_f, \mathbf{x}_e) = \mathbf{y}_f^\text{T} \mathbf{y}_e \tag{3}$$

According to (2), we see that the value of the scoring function is determined by the projection matrices $\boldsymbol{\theta} = \{\mathbf{W}_1, \mathbf{W}_2\}$.

The SPTM of (2) and (3) can be incoporated into the log-linear model for SMT (1) by introducing a new feature $h_{M+1}$ and a new feature weight $\lambda_{M+1}$. The new feature is defined as

$$h_{M+1}(F_i, E, A) = \sum_{(f,e) \in A} \text{sim}_{\boldsymbol{\theta}}(\mathbf{x}_f, \mathbf{x}_e). \tag{4}$$

---

[1] In our experiments, we compared dot product and the cosine similarity function and found that the former works better for nonlinear multi-layer neural networks, and the latter works better for linear neural networks. For the sake of clarity, we choose dot product when we describe the SPTM and its training in Sections 4 and 5, respectively.

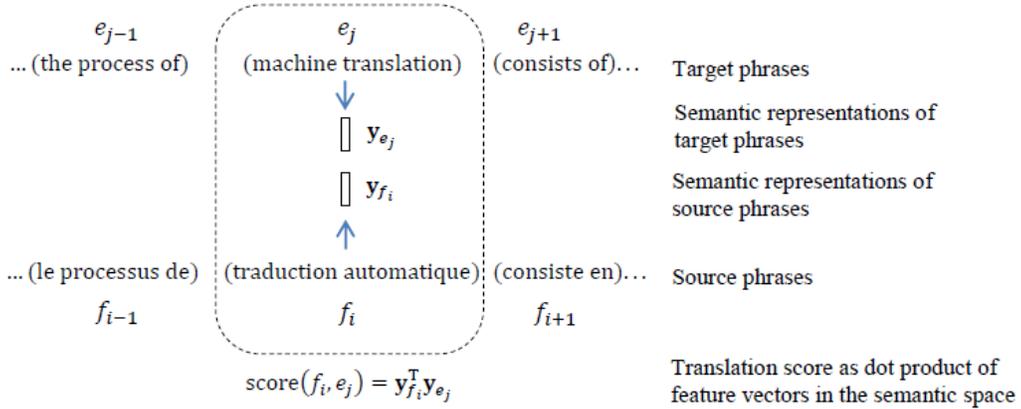

Figure 1. The architecture of the SPTM, where the mapping from a phrase to its semantic representation is shown in Figure 2.

Thus, the phrase-based SMT system, into which the SPTM is incorporated, is parameterized by $(\boldsymbol{\lambda}, \boldsymbol{\theta})$, where $\boldsymbol{\lambda}$ is a vector of a handful of parameters used in the log-linear model of (1), with one weight for each feature; and $\boldsymbol{\theta}$ is the projection matrices used in the SPTM defined by (2) and (3). In our experiments we take three steps to learn $(\boldsymbol{\lambda}, \boldsymbol{\theta})$:

1. Given a baseline phrase-based SMT system and a pre-set $\boldsymbol{\lambda}$ where $\lambda_{M+1} = 0$, we generate for each source sentence in training data an N-best list of translation hypotheses.
2. We fix $\boldsymbol{\lambda}$ and set $\lambda_{M+1} = 1$, and optimize $\boldsymbol{\theta}$ w.r.t. a loss function on training data.
3. We fix $\boldsymbol{\theta}$, and optimize $\boldsymbol{\lambda}$ using MERT [20] to maximize the BLEU score on development data.

In Section 5, we will describe Step 2 in detail because it is directly related to the SPTM training.

## 5. Training SPTM

This section describes the kind of loss function we employ with the SPTM and the algorithm to train the neural network weights using the loss function as the optimization objective.

We define the loss function $\mathcal{L}(\boldsymbol{\theta})$ as the negative of the N-best list based expected BLEU, denoted by xBleu$(\boldsymbol{\theta})$. In the reranking framework of SMT outlined in Section 3, xBleu$(\boldsymbol{\theta})$ over one training sample $(F_i, E_i)$ is defined as

$$\text{xBleu}(\boldsymbol{\theta}) = \sum_{E \in \text{GEN}(F_i)} P(E|F_i) \text{sBleu}(E_i, E) \quad (5)$$

where sBleu$(E_i, E)$ is the sentence-level BLEU score, and $P(E|F_i)$ is a normalized translation probability from $F_i$ to $E$ computed using *softmax* as

$$P(E|F_i) = \frac{\exp(\boldsymbol{\lambda}^\text{T} \mathbf{h}(F_i, E, A))}{\sum_{E \in \text{GEN}(F_i)} \exp(\boldsymbol{\lambda}^\text{T} \mathbf{h}(F_i, E, A))} \quad (6)$$

where $\boldsymbol{\lambda}^\text{T} \mathbf{h}$ is the log-linear model of (1), which also includes the feature derived from the SPTM as defined by (4).

Let $\mathcal{L}(\boldsymbol{\theta})$ be a loss function which is differentiable w.r.t. the parameters of the SPTM, $\boldsymbol{\theta}$. We can compute the gradient of the loss and learn $\boldsymbol{\theta}$ using gradient-based numerical optimization algorithms, such as L-BFGS or stochastic gradient descent (SGD).

### 5.1 Computing the Gradient

Since the loss does not explicitly depend on $\boldsymbol{\theta}$, we use the chain rule for differentiation:

$$\frac{\partial \mathcal{L}(\boldsymbol{\theta})}{\partial \boldsymbol{\theta}} = \sum_{(f,e)} \frac{\partial \mathcal{L}(\boldsymbol{\theta})}{\partial \text{sim}_{\boldsymbol{\theta}}(\mathbf{x}_f, \mathbf{x}_e)} \frac{\partial \text{sim}_{\boldsymbol{\theta}}(\mathbf{x}_f, \mathbf{x}_e)}{\partial \boldsymbol{\theta}} = \sum_{(f,e)} -\delta_{(f,e)} \frac{\partial \text{sim}_{\boldsymbol{\theta}}(\mathbf{x}_f, \mathbf{x}_e)}{\partial \boldsymbol{\theta}} \tag{7}$$

which takes the form of summation over all phrase pairs occurring either in a training sample (stochastic mode) or in the entire training data (batch mode). $\delta_{(f,e)}$ in (7) is known as the *error* term of the phrase pair $(f, e)$, and is defined as

$$\delta_{(f,e)} = -\frac{\partial \mathcal{L}(\boldsymbol{\theta})}{\partial \text{sim}_{\boldsymbol{\theta}}(\mathbf{x}_f, \mathbf{x}_e)} \tag{8}$$

It describes how the overall loss changes with the translation score of the phrase pair $(f, e)$. We will leave the derivation of $\delta_{(f,e)}$ to Section 5.1.2, and will first describe how the gradient of $\text{sim}_{\boldsymbol{\theta}}(\mathbf{x}_f, \mathbf{x}_e)$ w.r.t. $\boldsymbol{\theta}$ is computed.

### 5.1.1 Computing $\partial sim_{\boldsymbol{\theta}}(x_f, x_e)/\partial \boldsymbol{\theta}$

Without loss of generality, we use the following notations to describe a neural network:
- $\mathbf{W}_l$ is the projection matrix for the *l*-th layer of the neural network;
- $\mathbf{x}$ is the input word vector of a phrase;
- $\mathbf{z}^l = \mathbf{W}_l^T \mathbf{y}^{l-1}$ is the sum vector of the *l*-th layer; and
- $\mathbf{y}^l = \sigma(\mathbf{z}^l)$ is the output vector of the *l*-th layer, where $\sigma$ is an activation function;

Thus, the SPTM defined by (2) and (3) can be represented as

$$\mathbf{z}^1 = \mathbf{W}_1^T \mathbf{x}$$
$$\mathbf{y}^1 = \sigma(\mathbf{z}^1)$$
$$\mathbf{z}^2 = \mathbf{W}_2^T \mathbf{y}^1$$
$$\mathbf{y}^2 = \sigma(\mathbf{z}^2)$$
$$\text{sim}_{\boldsymbol{\theta}}(\mathbf{x}_f, \mathbf{x}_e) = (\mathbf{y}_f^2)^T \mathbf{y}_e^2$$

The gradient of the matrix $\mathbf{W}_2$ which projects the hidden vector to the output vector is computed as:

$$\frac{\partial \text{sim}_{\boldsymbol{\theta}}(\mathbf{x}_f, \mathbf{x}_e)}{\partial \mathbf{W}_2} = \frac{\partial (\mathbf{y}_f^2)^T}{\partial \mathbf{W}_2} \mathbf{y}_e^2 + (\mathbf{y}_f^2)^T \frac{\partial \mathbf{y}_e^2}{\partial \mathbf{W}_2} = \mathbf{y}_f^1 \left( \mathbf{y}_e^2 \circ \sigma'(\mathbf{z}_f^2) \right)^T + \mathbf{y}_e^1 \left( \mathbf{y}_f^2 \circ \sigma'(\mathbf{z}_e^2) \right)^T \tag{9}$$

where $\circ$ is the element-wise multiplication (Hadamard product). Applying the back propagation principle, the gradient of the projection matrix mapping the input vector to the hidden vector $\mathbf{W}_1$ is computed as

$$\frac{\partial \text{sim}_{\boldsymbol{\theta}}(\mathbf{x}_f, \mathbf{x}_e)}{\partial \mathbf{W}_1} = \mathbf{x}_f \left( \mathbf{W}_2 \left( \mathbf{y}_e^2 \circ \sigma'(\mathbf{z}_f^2) \right) \circ \sigma'(\mathbf{z}_f^1) \right)^T + \mathbf{x}_e \left( \mathbf{W}_2 \left( \mathbf{y}_f^2 \circ \sigma'(\mathbf{z}_e^2) \right) \circ \sigma'(\mathbf{z}_e^1) \right)^T \tag{10}$$

The derivation can be easily extended to a neural network with multiple hidden layers.

### 5.1.2 Computing $\delta_{(f,e)}$

To simplify the notation, we rewrite our loss function of (5) and (6) over one training sample as

$$\mathcal{L}(\boldsymbol{\theta}) = -\text{xBleu}(\boldsymbol{\theta}) = -\frac{G(\boldsymbol{\theta})}{Z(\boldsymbol{\theta})} \tag{11}$$

where

$$G(\boldsymbol{\theta}) = \sum_E \text{sBleu}(E, E_i) \exp(\boldsymbol{\lambda}^T \mathbf{h}(F_i, E, A))$$
$$Z(\boldsymbol{\theta}) = \sum_E \exp(\boldsymbol{\lambda}^T \mathbf{h}(F_i, E, A))$$

Combining (8) and (11), we have

$$\delta_{(f,e)} = \frac{\partial \text{xBleu}(\theta)}{\partial \text{sim}_\theta(\mathbf{x}_f, \mathbf{x}_e)} = \frac{1}{Z(\theta)} \left( \frac{\partial G(\theta)}{\partial \text{sim}_\theta(\mathbf{x}_f, \mathbf{x}_e)} - \frac{\partial Z(\theta)}{\partial \text{sim}_\theta(\mathbf{x}_f, \mathbf{x}_e)} \text{xBleu}(\theta) \right) \quad (12)$$

Because $\theta$ is only relevant to $h_{M+1}$ which is defined in (4), we have

$$\frac{\partial \boldsymbol{\lambda}^\text{T} \mathbf{h}(F_i, E, A)}{\partial \text{sim}_\theta(\mathbf{x}_f, \mathbf{x}_e)} = \lambda_{M+1} \frac{\partial h_{M+1}(F_i, E, A)}{\partial \text{sim}_\theta(\mathbf{x}_f, \mathbf{x}_e)} = \lambda_{M+1} N(f, e; A) \quad (13)$$

where $N(f, e; A)$ is the number of times the phrase pair $(f, e)$ occur in $A$. Combining (12) and (13), we end up with the following equation

$$\delta_{(f,e)} = \sum_{(E,A) \in GEN(F_i)} U(\theta, E) P(E|F_i) \lambda_{M+1} N(f, e; A) \quad (14)$$

where $U(\theta, E) = \text{sBleu}(E_i, E) - \text{xBleu}(\theta)$.

## 5.2 The Training Algorithm

In our experiments we train the parameters of the SPTM $\theta$ using the L-BFGS optimizer described in [1], together with the loss function described in (5). The gradient is computed as described in Sections 5.1. Even though the loss function is not convex, we found that the L-BFGS iterations over the complete training data (batch mode) minimizes the loss in practice in a desirable fashion; e.g., convergence of the algorithm was found to be smooth.

## 6. Experiments

We conducted our experiments on two Europarl translation tasks, English-to-French (EN-FR) and German-to-English (DE-EN). The data sets are published for the shared task in NAACL 2006 Workshop on Statistical Machine Translation (WMT06) [16]. For EN-FR, the training set contains 688K sentence pairs, with 21 words per sentence on average. The development set contains 2000 sentences. We used 2000 sentences from the WMT05 shared task as the first test set (TEST1), and the 2000 sentences from the WMT06 shared task as the second test set (TEST2). For DE-EN, the training set contains 751K sentence pairs, with 21 words per sentence on average. The official development set used for the shared task contains 2000 sentences. We used 2000 sentences from the WMT05 shared task as TEST1, and the 2000 sentences from the WMT06 shared task as TEST2. We used the Moses system [15] as our baseline phrase-based SMT system. We built two baseline systems, each for one language pair. The metric used for evaluation is case insensitive BLEU score [21]. We also performed a significance test using the paired *t*-test. Differences are considered statistically significant when the *p*-value is less than 0.05.

### 6.1 Results

Table 2 shows the main results measured in BLEU evaluated on TEST1 and TEST2, where Row 1 is the baseline system. Rows 2 to 5 are the systems enhanced by integrating different versions of the SPTM. **SPTM** in Row 2 is the model described in Sections 4. As illustrated in Figure 2, the number of the nodes in the input layer is the vocabulary size $d$. Both the hidden layer and the output layer have 100 nodes. That is, $\mathbf{W}^1$ is a $d \times 100$ matrix and $\mathbf{W}^2$ a $100 \times 100$ matrix. Table 2 shows that **SPTM** leads to a substantial improvement over the baseline system across all test sets, with a statistically significant margin from 0.7 to 1.0 BLEU points.

We have developed a set of variants of **SPTM**, as shown in Rows 3 to 5, to investigate two design choices we made in developing the SPTM: (1) whether to use a linear projection or a multi-layer nonlinear projection; and (2) whether to compute the phrase similarity using word-word similarities as suggested by e.g., the lexical weighting model [17].

**SPTM**$_L$ (Row 3) uses a linear neural network to project a word vector of a phrase $\mathbf{x}$ to a feature vector $\mathbf{y} \equiv \phi(\mathbf{x}) = \mathbf{W}^\text{T} \mathbf{x}$, where $\mathbf{W}$ is a $d \times 100$ projection matrix. The translation score of a source phrase *f* and a target phrase *e* is measured as the similarity of their feature vectors. We choose cosine similarity because it works better than dot product for linear projection. **SPTM$_W$** (Row 4) computes the phrase similarity using word-word similarity scores. This follows the common smoothing strategy of addressing the data sparseness problem in modeling phrase translations, such as the lexical weighting model [17] and the word factored n-gram translation model [24].

| # | Systems | EN-FR | | DE-EN | |
|---|---|---|---|---|---|
| | | TEST1 | TEST2 | TEST1 | TEST2 |
| 1 | Baseline | 32.79 | 32.84 | 26.04 | 26.04 |
| 2 | SPTM | **33.79$^\alpha$** | **33.81$^\alpha$** | **26.82$^\alpha$** | **26.72$^\alpha$** |
| 3 | SPTM$_L$ | 33.56$^{\alpha\beta}$ | 33.51$^{\alpha\beta}$ | 26.67$^\alpha$ | 26.50$^{\alpha\beta}$ |
| 4 | SPTM$_W$ | 33.21$^{\alpha\beta}$ | 33.27$^{\alpha\beta}$ | 26.56$^{\alpha\beta}$ | 26.49$^{\alpha\beta}$ |
| 5 | SPTM$_{L-W}$ | 33.25$^{\alpha\beta}$ | 33.35$^{\alpha\beta}$ | 26.46$^{\alpha\beta}$ | 26.33$^{\alpha\beta}$ |
| 6 | 2 + 4 | 33.79$^\alpha$ | 33.81$^\alpha$ | 26.81$^\alpha$ | 26.73$^\alpha$ |
| 7 | BLTM$_{PR}$ | 32.78$^\beta$ | 32.95 | 26.06$^\beta$ | 26.09$^\beta$ |
| 8 | DPM | 32.90$^\beta$ | 32.99$^{\alpha\beta}$ | 26.20$^{\alpha\beta}$ | 26.16$^\beta$ |

Table 2: Main results (BLEU scores) of semantic-based phrase translation models. The superscripts $\alpha$ and $\beta$ indicate statistically significant difference ($p < 0.05$) from **Baseline** and **SPTM**, respectively.

Two observations can be made by comparing **SPTM** in Row 2 to its variants in Rows 3-5. First of all, it is more effective to model the phrase translation directly than decomposing it into word-word translations in the SPTMs (Row 2 vs. Row 4 and Row 3 vs. Row 5). Moreover, unlike the case of traditional phrase translation models, combining the phrase model and the word model does not lead to any visible improvement (Row 6 vs. Row 2), indicating that with semantic representations, a phrase model is no longer sparser than a word model. Second, we see that in phrase models (Rows 2 and 3) the nonlinear projection is able to capture more sophisticated semantic information and leads to better results than the linear projection.

## 6.2 Comparing with Previous Latent Semantic Models

This section compares the best version of the SPTM i.e., **SPTM** in Row 2 of Table 2, with two state-of-the-art latent semantic models that are originally trained on clicked query-document pairs (i.e., clickthrough data extracted from search logs) for query-document matching [9]. To adopt these models for SMT, we view source-target sentence pairs as clicked query-document pairs, and trained both models using the same methods as in [9] on the parallel bilingual training data described earlier.

The results are shown in Table 2. **BTLM$_{PR}$** (Row 7) is an extension to PLSA, and is the best performer among different versions of the Bi-Lingual Topic Model (BLTM) described in [9]. BLTM with Posterior Regularization (**BLTM$_{PR}$**) is trained on parallel training data using the EM algorithm with a constraint enforcing a source sentence and its paralleled target sentence to not only share the same prior topic distribution, but to also have similar fractions of words assigned to each topic. We incorporated the model into the log-linear model for SMT (1) as follows. First of all, the topic distribution (i.e., semantic representation) of a source sentence $F_i$, denoted by $P(z|F_i)$, is induced from the learned topic-word distributions using EM. Then, each translation candidate $E$ in the N-best list GEN($F_i$) is scored as

$$P(E|F_i) = \prod_{w \in E} \sum_z P(w|z)P(z|F_i)$$

$P(F_i|E)$ can be similarly computed. Finally, the logarithms of the two probabilities are incorporated into the log-linear model of (1) as two additional features.

**DPM** (Row 8) is the Discriminative Projection Model described in [9]. **DPM** uses a matrix **W** to project a word vector of a sentence to a feature vector. **W** is trained on parallel training data using a Siamese neural network approach, S2Net [26], as follows. For each source sentence in training data, we treat it and its paralleled translation in target language as a positive pair, and we randomly selected 4 other target sentences from training data to form 4 negative pairs. **W** is trained in such a way that a positive source-target sentence pair has a higher similarity (i.e., cosine similarity) than that of the negative ones of the same source sentence. **DPM** can be incorporated into the log-linear model for SMT (1) by introducing a new feature $h_{M+1}$. Let **x** be the word vector of a source sentence $F_i$ (or its translation candidate $E$), and **y** be the projected feature vector, i.e., $\mathbf{y} = \mathbf{W}^T\mathbf{x}$. The new feature is defined as

$$h_{M+1}(F_i, E) \equiv \text{sim}_{\mathbf{W}}(\mathbf{x}_{F_i}, \mathbf{x}_E) = \frac{\mathbf{y}_{F_i}^T \mathbf{y}_E}{\|\mathbf{y}_{F_i}\|\|\mathbf{y}_E\|}$$

Similar to that BLTM is an extension to PLSA, DPM can be viewed as an extension of LSA where bilingual parallel data can be explored for translation model training. As we see from Table 2, both latent semantic models, although leading to some slight improvement over **Baseline**, are much less effective than **SPTM** which is based on a multi-layer neural network trained on the N-best lists using a loss function that tailors to the BLEU metric. However, we found in our experiments that these models can be useful for "pre-training" to provide a good initial model that not only speeds up the SPTM training but also leads to a better final model.

### 6.3 Discussion

Although SGD has been advocated for neural network training due to its simplicity and its robustness to local minimum, we found that in our task the L-BFGS based batch training performs well despite the non-convexity in our loss. Another merit of batch training is that the gradient over all training data can be computed efficiently. As shown in Section 5, computing $\partial \text{sim}_\theta(\mathbf{x}_f, \mathbf{x}_e)/\partial \theta$ requires large-scale matrix multiplications, and is expensive for multi-layer neural networks. Eq. (7) suggests that $\partial \text{sim}_\theta(\mathbf{x}_f, \mathbf{x}_e)/\partial \theta$ and $\delta_{(f,e)}$ can be computed separately, thus making the computation cost of the former term only depends on the number of phrase pairs in the phrase table, but not the size of training data. Therefore, the training method described in Section 5 can be used on larger amounts of training data with little difficulty.

### 7. Conclusions

The work presented in this paper makes two important contributions. First, we develop a novel phrase translation model for SMT, where the translation score of a pair of source-target phrases is represented as the distance between their feature vectors in a low-dimensional, continuous-valued semantic space. The semantic space is derived from the representations generated using a multi-layer neural network. Second, we present a new learning method to train the weights in the multi-layer neural network for the end-to-end BLEU metric directly. The training method is based on L-BFGS. We describe in detail how the gradient in closed form, as required for efficient optimization, is derived. The objective function, which takes the form of the expected BLEU computed from N-best lists, is very different from the usual objective functions used in most existing neural networks, e.g., cross entropy or mean square error [11, 12]. We hence have provided details in the derivation of the gradient, which can serve as an example to guide the derivation of neural network learning with other non-standard objective functions in the future.

Our evaluation on two Europal translation tasks show that incorporating the SPTM into the log-linear framework of SMT significantly improves the performance of a state-of-the-art phrase-based SMT system, leading to a gain between 0.7 to 1.0 BLEU points. Careful implementation of the L-BFGS optimization based on the BLEU-centric objective function, together with the associated closed-form gradient, is a key to the success.

A natural extension of this work is to expand the model and learning algorithm from shallow to deep neural networks. The deep models are expected to produce more powerful and flexible semantic representations, and thus greater performance gain than what is presented in this paper.